\PassOptionsToPackage{dvipsnames}{xcolor}
\documentclass[sn-mathphys-num]{sn-jnl}
\newcommand{\keypoint}[1]{\vspace{0.1cm}\noindent\textbf{#1}\;}
\usepackage{amsmath}
\usepackage{amsfonts}
\usepackage{booktabs} 
\usepackage{color}
\usepackage{colortbl}
\usepackage{graphicx}
\usepackage{multirow}
\usepackage{amsmath}
\usepackage{amssymb}
\usepackage{mathtools}
\usepackage{xcolor}
\usepackage{wrapfig}
\usepackage{lipsum}
\usepackage{lmodern}
\usepackage{xcolor}
\usepackage{hyperref}
\usepackage[linesnumbered,ruled]{algorithm2e}
\usepackage{algpseudocode}
\usepackage{algorithmicx}
\usepackage{algcompatible}
\definecolor{commentcolor}{RGB}{110,154,155}

\usepackage{xcolor,pifont}
\newcommand*\colourcheck[1]{%
  \expandafter\newcommand\csname #1check\endcsname{\textcolor{#1}{\ding{52}}}%
}
\newcommand*\colourcross[1]{%
  \expandafter\newcommand\csname #1cross\endcsname{\textcolor{#1}{\ding{55}}}%
}
\colourcheck{blue}
\colourcheck{green}
\colourcheck{black}
\colourcheck{red}
\colourcross{red}
\colourcheck{Maroon}
\colourcross{black}
\colourcross{green}
\colourcross{Green}
\colourcross{MidnightBlue}

\usepackage[capitalize]{cleveref}
\crefname{section}{Sec.}{Secs.}
\Crefname{section}{Section}{Sections}
\Crefname{table}{Table}{Tables}
\crefname{table}{Tab.}{Tabs.}
\crefname{algorithm}{Algo.}{Algos.}

\definecolor{deepGreen}{RGB}{0,153,0}
\definecolor{orange}{RGB}{255,125,0}
\def\red#1{\textcolor[rgb]{1,0,0}{#1}}
\def\blue#1{\textcolor[rgb]{0,0,1}{#1}}

\definecolor{Gray}{gray}{0.9}
\definecolor{pink}{RGB}{255, 234, 232}
\definecolor{good}{RGB}{214, 232, 212}
\definecolor{reasonable}{RGB}{188, 200, 211}
\definecolor{abstract}{RGB}{242, 214, 213}
\definecolor{columbiablue}{rgb}{0.61, 0.87, 1.0}

\def\eg{\emph{e.g.}}

\def\ie{\emph{i.e.}}

\begin{document}

\title[Article Title]{Training-free Temporal Object Tracking in Surgical Videos}

\author[1]{\fnm{Subhadeep} \sur{Koley}}\email{subhadeep.koley@medtronic.com}

\author[1]{\fnm{Abdolrahim} \sur{Kadkhodamohammadi}}\email{rahim.mohammadi@medtronic.com}

\author[1]{\fnm{Santiago} \sur{Barbarisi}}\email{santiago.barbarisi@medtronic.com}

\author[1,2]{\fnm{Danail} \sur{Stoyanov}}\email{danail.stoyanov@medtronic.com}

\author[1]{\fnm{Imanol} \sur{Luengo}}\email{imanol.luengo@medtronic.com}

\affil[1]{\orgname{Medtronic plc.}, \orgaddress{\city{London}, \country{UK}}}
\affil[2]{\orgname{UCL Hawkes Institute, University College London}, \orgaddress{\city{London}, \country{UK}}}

\abstract{\textbf{Purpose:} In this paper, we present a novel approach for online object tracking in laparoscopic cholecystectomy (LC) surgical videos, targeting localisation and tracking of critical anatomical structures and instruments. Our method addresses the challenges of costly pixel-level annotations and label inconsistencies inherent in existing datasets.

\textbf{Methods:} Leveraging the inherent object localisation capabilities of pre-trained text-to-image diffusion models, we extract representative features from surgical frames without any training or fine-tuning. Our tracking framework uses these features, along with cross-frame interactions via an affinity matrix inspired by query-key-value attention, to ensure temporal continuity in the tracking process.

\textbf{Results:} Through a pilot study, we first demonstrate that diffusion features exhibit superior object localisation and consistent semantics across different decoder levels and temporal frames. Later, we perform extensive experiments to validate the effectiveness of our approach, showcasing its superiority over competitors for the task of temporal object tracking. Specifically, we achieve a per-pixel classification accuracy of $79.19\%$, mean Jaccard Score of $56.20\%$, and mean F-Score of $79.48\%$ on the publicly available CholeSeg8K dataset.

\textbf{Conclusion:} Our work not only introduces a novel application of text-to-image diffusion models but also contributes to advancing the field of surgical video analysis, offering a promising avenue for accurate and cost-effective temporal object tracking in minimally invasive surgery videos.}


\maketitle

\section{Introduction}
\label{sec:intro}
Surgical video analysis has now been proven to be a crucial tool for providing surgeons with useful analytics and insights on minimally invasive surgeries \cite{grammatikopoulou2024spatio, chadebecq2023artificial, chadebecq2020computer}. Among them temporal object tracking deals with tracking segmentation masks of critical instruments and anatomies present in a surgical frame along the temporal direction. It aids computer-assisted intervention (CAI) in three ways. \textit{(i) Pre-operative surgeon training:} prepares surgeons in correctly identifying and segregating critical organ structure and surgery-planning. \textit{(ii) Intra-operative guidance:} provides \textit{real-time} feedback to surgeons during operation, helping them establish Critical View of Safety (CVS) to minimise the risk of unwanted organ injury \cite{hong2020cholecseg8k}. Finally, \textit{(iii) Post-operative case study:} equips medical practitioners with analytics to study failure and success cases \cite{chen2023surgnet}.

In this paper, we propose a {training-free} framework for temporal object tracking in {Laparoscopic Cholecystectomy} (LC) surgery videos, that deals with surgical extraction of the gallbladder. The accurate identification of certain anatomies like cystic duct and artery is critical to avoid the risk of bile duct injury, a serious injury associated with severe post-operative morbidity, lower long-term survival and negative impact on patient's quality of life \cite{hong2020cholecseg8k}. Temporal object tracking can help avoid these scenarios by providing accurate localisation of crucial structures throughout the entire video.

Temporal object tracking in surgical videos is however non-trivial. {Firstly}, the high annotation cost of pixel-level masks makes large-scale {fully-supervised} training of tracking networks (with mask-frame pairs) unfeasible \cite{grammatikopoulou2024spatio}. {Second}, producing accurate annotations is even more difficult for surgical videos, where the community not only faces data scarcity but also requires medical expertise for accurate annotation \cite{grammatikopoulou2024spatio}. Moreover, most of the existing datasets \cite{hong2020cholecseg8k}, were annotated with {semi-automated} segmentation pipeline, which inherently introduces label inconsistencies, to some extent. Consequently, fully supervised training might be erroneous in such scenarios. 

\begin{figure}[!htbp]
    \centering
    \includegraphics[width=0.9\textwidth]{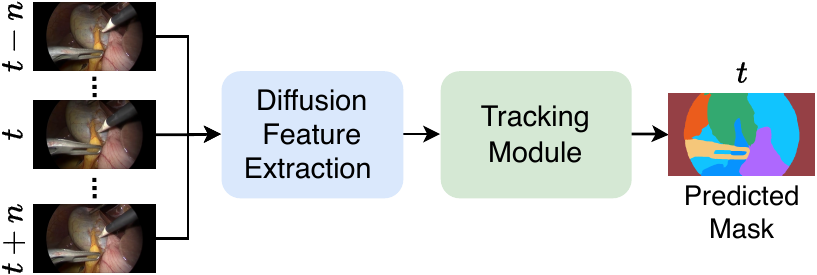}
    \caption{Our method uses pre-trained text-to-image diffusion model \cite{rombach2022high} to extract features from surgical frames (Sec. \ref{sec:diff_feat}). Our tracking module (Sec. \ref{sec:seg}) uses these features with cross-frame interactions via an affinity matrix, to predict masks in the temporal direction for the entire clip.}
    \label{fig:teaser}
\end{figure}

Apart from high-resolution text-to-image generation \cite{zhang2023adding}, diffusion models also depicts exceptional performance in semantic local editing \cite{ruiz2023dreambooth}, object detection \cite{xu20233difftection}, correspondence learning \cite{tang2023emergent}, among different tasks. Therefore, we posit (and validate) that its internal representations {inherently} encompass some form of object-{localisation} and {grouping}, even though not {specifically} aimed during training. 

In this paper, we dig into the internal representations of {pre-trained} text-to-image latent diffusion models \cite{rombach2022high} to extract representative features from surgical frames that contains useful geometry and object-localisation. We then use these features for temporal object tracking in surgical videos \textit{without} any sort of training or fine-tuning. Thus, alleviating the issues of pixel-level annotation cost and noisy ground truths.

Our answer to this seemingly complex challenge of accurate object tracking is simple yet effective (\cref{fig:teaser}). We first  (\cref{sec:pilot}) validate that internal features of a pre-trained diffusion model \cite{rombach2022high} inherently holds superior object localisation, making them suitable for temporal tracking. We find these features to showcase different feature-granularity based on different decoder levels of the architecture. Moreover, we also discover that the semantics of these features enforce temporal-consistency. The rest of the framework revolves around identifying the most optimal layer, timestep, and other network parameters tailored to our task of object tracking. Our object tracking module uses these features and the first frame GT masks (provided by the user) to automatically track those masks temporally across the \textit{entire} video. Crucially, in the tracking module, we utilise cross-frame interactions in the temporal direction, inspired by {query-key-value} attention \cite{vaswani2017attention}. Particularly, we calculate an affinity matrix using the diffusion features of consecutive frames. This affinity matrix is then multiplied by the first frame GT mask, to produces the next frame mask. Moreover, we incorporate a limited history of previous predictions in current mask calculation, to maintain temporal-consistency.

In summary, we uncover the latent potential of text-to-image diffusion models as an efficient online mask tracker in surgical videos. In addition, we leverage the cross-frame interactions via an affinity matrix while involving multi-frame features during mask prediction to maintain temporal-continuity. Lastly, extensive experiments and ablative studies demonstrates the proposed method to be the best performer among all training-free competitors.

\section{Related Work}
\label{sec:related}
\keypoint{Surgical Video Object Tracking.}
It deals with the automatic tracking of {anatomical structures} (\eg, liver, gallbladder) and {surgical instruments} (\eg, grasper, electrocautery) masks in surgical videos. Due to the nature of the task and unavailability of {mask-annotated surgical video} datasets, this field remains relatively under-explored. Most of the works typically use different backbones like HRNet32 \cite{sun2019high} or UNet \cite{ronneberger2015u} to perform segmentation individually on each frame of the surgical videos. While SP-TCN \cite{grammatikopoulou2024spatio} trains a spatio-temporal decoder on top of encoder features, MSPH \cite{zhang2024towards} utilises class-labels to perform segmentation. Among recent methods, MFC \cite{zhao2023masked} resorts to domain-adaptation, while SurgNet \cite{chen2023surgnet} uses self-supervised pre-training. However, most of these methods require some form of training or fine-tuning with mask-annotations \cite{grammatikopoulou2024spatio} or weak-labels \cite{zhang2024towards}. Nonetheless, the high annotation cost of pixel-level masks renders large-scale mask-frame pair generation impractical. Considering the current status quo within the medical vision community, it still concerns gathering enough video frames of surgical procedures, let alone pixel-perfect paired masks.

\keypoint{Diffusion Models for Visual Tasks.}
Diffusion models have now been established to be the gold-standard for high-resolution 2D and 3D image- and video-generation frameworks like Stable Diffusion \cite{rombach2022high}, ControlNet \cite{zhang2023adding}, or VideoCrafter2 \cite{chen2024videocrafter2}. Beyond generation, diffusion-based models (\eg, Dreambooth \cite{ruiz2023dreambooth}) are also being used for photorealistic image-editing. Besides generative tasks, it is also being used in other downstream discriminative tasks like recognition \cite{li2023your}, bounding-box tracking \cite{luo2024diffusiontrack}, object-detection \cite{xu20233difftection} or semantic correspondence learning \cite{tang2023emergent, hedlin2023unsupervised, zhang2023tale}, etc. In this paper, our goal is to delve deeper into pre-trained SD's internal representations to find out its efficacy in temporal object tracking in laparoscopic videos {without} training. Notably, SD was pre-trained on high-quality natural image-text pairs from LAION-5B dataset and \textit{no surgical image} was used \cite{rombach2022high}.

\section{Background: Diffusion Model}
\label{sec:diffusion}

Diffusion models generate images by iterative elimination of noise from a 2D isotropic Gaussian noise image \cite{rombach2022high}. It consists of two reciprocal processes -- forward and reverse diffusion \cite{rombach2022high}. The \textit{forward} process adds random Gaussian noise to a clean image $\mathbf{x}_0 \in \mathbb{R}^{h\times w\times 3}$ from the training dataset in an iterative manner (for {$t$} timesteps) to create a noisy image $\mathbf{x}_t \in \mathbb{R}^{h\times w\times 3}$. During the \textit{reverse} process, a denoising UNet \cite{ronneberger2015u} $\mathcal{U}_\theta$ is trained (with an $l_2$ objective), which estimates the input noise $\epsilon \approx \mathcal{U}_\theta(\mathbf{x}_t,t)$ from the noisy image $\mathbf{x}_t$ at each $t$. Once trained, $\mathcal{U}_\theta$ can recover the original image from a noisy image. The inference procedure starts from a random $2D$ noise $\mathbf{x}_T$$\sim$$\mathcal{N}(0,\mathbf{I})$. The well-trained $\mathcal{U}_\theta$ is employed iteratively (for $T$ timesteps) to remove noise from each timestep progressively to get a cleaner image $\mathbf{x}_{t-1}$. This eventually generates one of the cleanest samples $\mathbf{x}_0$ from the target distribution.

\subsection{Latent Diffusion Model (LDM)}
\label{sec:ldm}
Training and inference of vanilla diffusion model is time-consuming as it operates on the full image resolution (\ie, $\mathbf{x}_0 \in \mathbb{R}^{h\times w\times 3}$). Contrarily, in Latent Diffusion Model \cite{rombach2022high} (\ie, Stable Diffusion -- SD), denoising occurs on the encoder latent space, making it much faster and stable. In the {first} stage, SD trains a variational autoencoder (an encoder $\mathcal{E}(\cdot)$ and a decoder $\mathcal{D}(\cdot)$ in sequence). $\mathcal{E}(\cdot)$ converts the input image into its latent representation $\mathbf{z}_0 =\mathcal{E}(\mathbf{x}_0) \in \mathbb{R}^{\frac{h}{8}\times \frac{w}{8}\times d}$. In the {second} stage, SD trains a UNet \cite{ronneberger2015u} $\mathcal{U}_\theta$, that performs denoising directly on the latent images. $\mathcal{U}_\theta$ comprises $12$ encoding, $1$ bottleneck, and $12$ decoding blocks \cite{rombach2022high}. Inside these encoding and decoding layers, there are $4$ downsampling $(\mathcal{U}_{\mathbf{d}}^{1-4})$ and $4$ upsampling $(\mathcal{U}_{\mathbf{u}}^{1-4})$ layers respectively. A pre-trained CLIP language encoder \cite{radford2021learning} $\mathcal{T}(\cdot)$ converts the textual prompt $\mathbf{c}$ into token-sequence, that governs $\mathcal{U}_\theta$ through cross-attention. $\mathcal{U}_\theta$ is trained over an $l_2$ objective as: $\mathcal{L}_\text{SD}=\mathbb{E}_{\mathbf{z}_t,t,\mathbf{c},\epsilon} ({||\epsilon-\mathcal{U}_{\theta}(\mathbf{z}_t,t,\mathcal{T}(\mathbf{c}))||}_2^2)$. During testing, a noisy latent $\mathbf{z}_t$ is sampled directly as: $\mathbf{z}_t$$\sim$$\mathcal{N}(0,\mathbf{I})$. $\mathcal{U}_{\theta}$ removes noise from $\mathbf{z}_t$ iteratively over $T$ timesteps (conditioned on $\mathbf{c}$) to yield a denoised latent image $\Hat{\mathbf{z}}_0\in \mathbb{R}^{\frac{h}{8}\times \frac{w}{8}\times d}$. The final image is generated as: $\Hat{\mathbf{x}}=\mathcal{D}(\Hat{\mathbf{z}}_0)\in \mathbb{R}^{h\times w\times 3}$.

\subsection{Pilot Study: Analysing SD Internal Features}
\label{sec:pilot}

Apart from high-resolution text-to-image generation \cite{rombach2022high, zhang2023adding}, SD also delineates excellent performance in semantic local editing \cite{ruiz2023dreambooth}, object detection \cite{xu20233difftection} and correspondence learning \cite{tang2023emergent}. Therefore, it is reasonable to assume that its internal representations implicitly hold some form of object-{localisation} and {grouping}, even though not explicitly aimed during training. We hypothesise that, \textit{(i)} internal feature maps of SD's \cite{rombach2022high} UNet decoder contain highly localised \textit{hierarchical object groupings}; \textit{(ii)} the semantics of this localised information is \textit{temporally-consistent}. 

\begin{figure}[!htbp]
    \centering
    \includegraphics[width=1\linewidth]{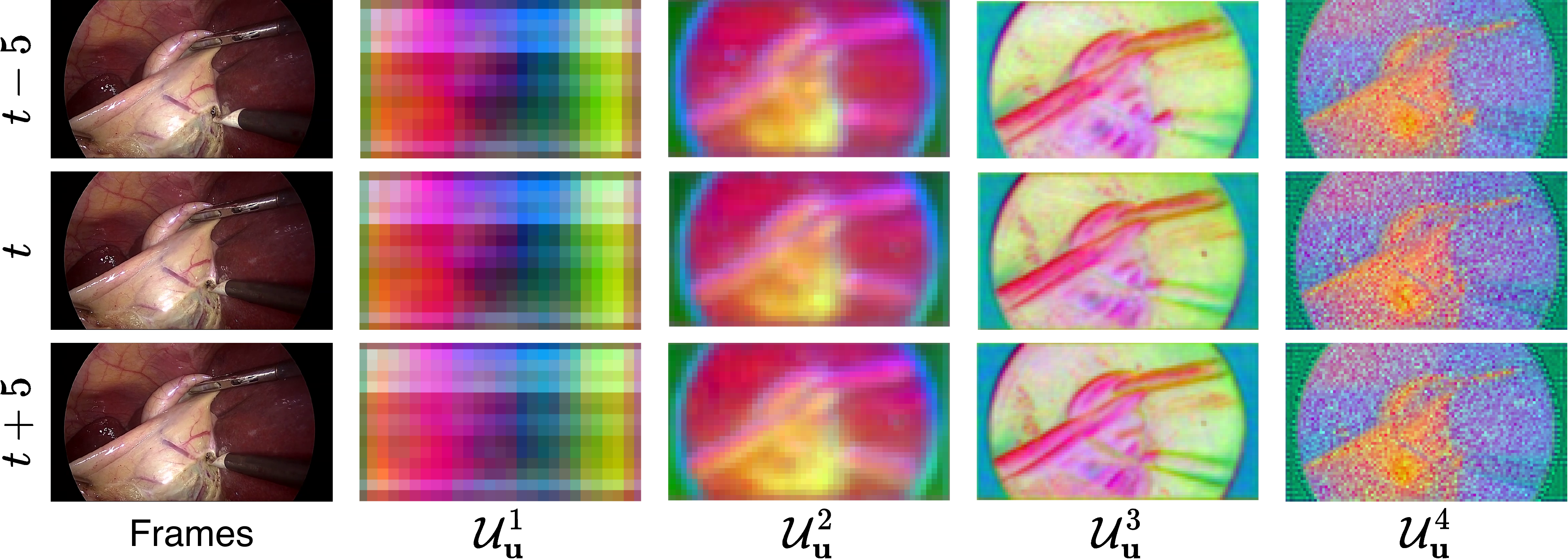}
    \caption{PCA rendering of SD's $\mathcal{U}_{\mathbf{u}}^{n}$ features from different levels ($n=$$\{1,2,3,4\}$) and time-frames. Notably, same anatomy/instrument is represented by same colour across temporal direction.}
    \label{fig:pilot}
\end{figure}

To verify this, we extract SD UNet decoder's internal features (from $\mathcal{U}_{\mathbf{u}}^{1-4}$) of surgical video frames with timestep $t=200$ and a {null-prompt} (\ie, $\mathtt{``~"}$). We perform Principal Component Analysis (PCA) on the extracted features and plot the top-$3$ principal components as RGB image (\cref{fig:pilot}). A few observations in \cref{fig:pilot} validate our hypothesis -- \textit{(i) object grouping:} instrument and anatomy features in the PCA maps are highly {localised} and {distinct}; \textit{(ii) feature-granularity:} features extracted from different decoder levels depict different feature-granularity. Features shift from coarse (\ie, low-frequency) to fine-grained (\ie, high-frequency) with increasing decoder levels; \textit{(iii) temporal-consistency:} SD internal features are temporally-consistent (\ie, same anatomy/instrument is represented by same colour across temporal direction). This resolves two primary bottlenecks of temporal object tracking -- \ie, temporal-consistency and object-localisation without GT masks.

\section{Methods}
\label{sec:method}
\subsection{Overview}
We aim to perform temporal object tracking of surgical videos without explicit training or fine-tuning, therefore alleviating the requirement of costly pixel-level mask annotation. Consequently, given the first frame GT mask $m_1$ from a surgical video containing $\{i_i\}_{i=1}^N$ frames, our method can track all individual structures present in $m_1$ on the rest $N-1$ frames to generate the corresponding masks $\{m_2, m_3, m_4, \cdots, m_N\}$.

\begin{figure}[!htbp]
    \centering
    \includegraphics[width=1\linewidth]{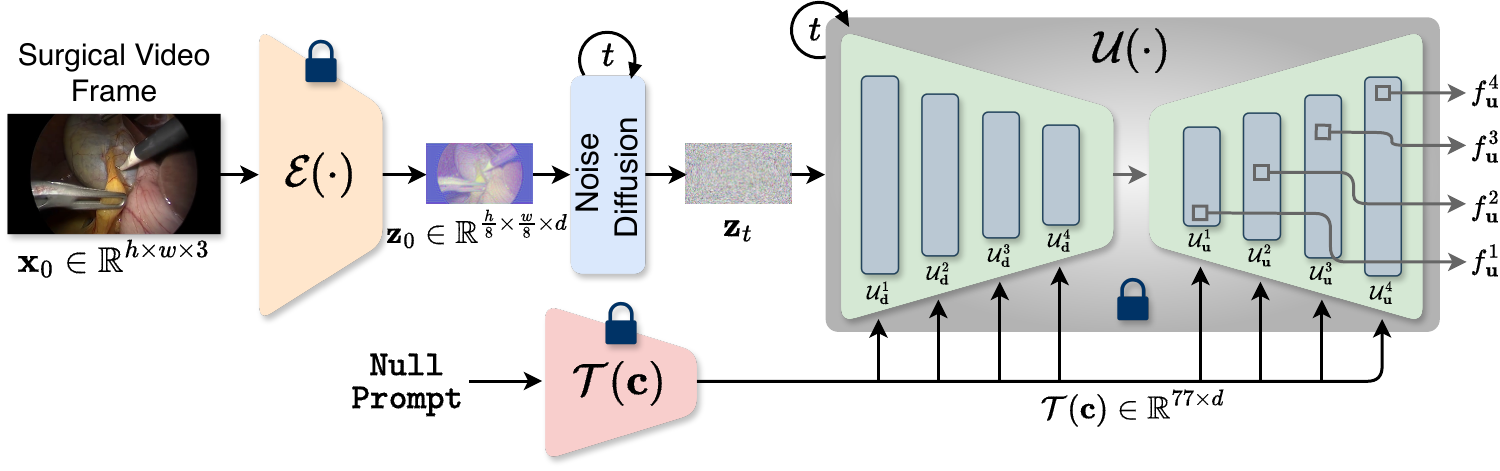}
    \caption{Proposed diffusion feature extraction pipeline. Given a surgical frame $\mathbf{x}_0$, it is passed through the encoder $\mathcal{E}(\cdot)$ to generate a latent representation $\mathbf{z}_0$, followed by forward diffusion to produce a noisy latent $\mathbf{z}_t$. The null prompt $\mathcal{T}(\mathbf{c})$ and noisy latent representation $\mathbf{z}_t$ are then passed through the UNet $\mathcal{U}(\cdot)$ to extract features from its internal decoders (Sec. \ref{sec:diff_feat}).}
    \label{fig:arch}
\end{figure}

We introduce three salient designs. \textit{Firstly}, we leverage the information-rich interim representation of text-to-image SD model for training free temporal object tracking. \textit{Second}, we use the interplay between the subsequent video frames with an affinity matrix. \textit{Finally}, we incorporate features from $n$ previous frames to ensure temporal-consistency.

\subsection{Stable Diffusion as Backbone Feature Extractor}
\label{sec:diff_feat}
We intend to use SD as a backbone feature extractor for temporal object tracking. Efficient utilisation of the SD internal features is however tricky and depends on a few {application-specific} hyperparameters \cite{tang2023emergent}. The extracted internal feature maps of SD vary significantly for different {diffusion timesteps} and {UNet architecture levels}. Thus, an optimum interplay between them is necessary to utilise the most of it. Given an image-prompt pair $\{\mathbf{x}, \mathbf{c}\}$, we generate the latent representation of $\mathbf{x}\in\mathbb{R}^{h\times w\times 3}$ as $\mathbf{z}_0 =\mathcal{E}(\mathbf{x}) \in \mathbb{R}^{\frac{h}{8}\times \frac{w}{8}\times d}$. With timestep $t$, Gaussian noise is added on $\mathbf{z}_0$ using a noise scheduler to transform it to the $t^{th}$ step noisy latent $\mathbf{z}_t$. The prompt embedding (\cref{sec:ldm}) $\mathcal{T}(\mathbf{c})$, $\mathbf{z}_t$, and $t$ are passed to the denoising UNet \cite{rombach2022high} $\mathcal{U}(\cdot)$ to extract the internal features from its decoders $\mathcal{U}_{\mathbf{u}}^{n} \rightarrow f_{\mathbf{u}}^{n}$. For example, in SD v2.1, an input frame of $\mathbf{x}$ of size $\mathbb{R}^{h\times w\times 3}$ would produce features $f_{\mathbf{u}}^{1}\in \mathbb{R}^{\frac{h}{32}\times \frac{w}{32}\times 1280}$, $f_{\mathbf{u}}^{2}\in \mathbb{R}^{\frac{h}{16}\times \frac{w}{16}\times 1280}$, $f_{\mathbf{u}}^{3}\in \mathbb{R}^{\frac{h}{8}\times \frac{w}{8}\times 640}$, and $f_{\mathbf{u}}^{4}\in \mathbb{R}^{\frac{h}{8}\times \frac{w}{8}\times 320}$ from $\mathcal{U}_{\mathbf{u}}^{1}$, $\mathcal{U}_{\mathbf{u}}^{2}$, $\mathcal{U}_{\mathbf{u}}^{3}$, and $\mathcal{U}_{\mathbf{u}}^{4}$ respectively. As our dataset \cite{hong2020cholecseg8k} does not contain paired text prompts, we use null-prompts (\ie, $\mathtt{``~"}$) in place of $\mathbf{c}$. The feature extraction procedure is depicted in \cref{fig:arch}.

\subsection{Temporal Tracking Module}
\label{sec:seg}

Utilising the extracted SD features for temporal object tracking is however non-trivial. It revolves around two key aspects. Firstly, selecting optimal SD feature representative enough for tracking the seemingly intricate instruments and anatomies present in surgical videos. Second, maintaining temporal-consistency between the segmented frames while doing so. In addressing the former, we perform a thorough ablative study (\cref{sec:abal}) to determine the best set of parameters. In addressing the latter, we leverage the \textit{cross-frame interactions} via an affinity matrix while involving multi-frame features during mask prediction to maintain temporal-continuity. 

\begin{center}
\begin{minipage}{0.7\textwidth}
\begin{algorithm}[H]
\small
\label{algo:seg}
\DontPrintSemicolon
\caption{Temporal Mask Tracking}
\textbf{Input:} First image-mask pair: $\{i_1, m_1\}$, Rest of the frames: $\{i_2, \cdots i_N\}$, Masking window: $n$, Temperature: $\tau$

\textbf{Output:} Tracked masks: $\{m_2, \cdots, m_N\}$

\medskip

$f_{\text{list}}^{2} \longleftarrow \{\}$

\For{$1 \leq i \leq$ N}{
    $f_{\text{list}}^{2} \longleftarrow \{f_{\text{list}}^{2}, \mathcal{U}_{\mathbf{d}}^{2}(i_i)\}$

    }

$\mathcal{N} \longleftarrow \mathtt{SpatialMask}(n)$

$m_{\text{list}} \longleftarrow \{\}$

\For{$2 \leq i \leq$ N}{
    $f_{i}^{2} \longleftarrow f_{\text{list}}^{2}[i_i]$

    $f_{i-1}^{2} \longleftarrow f_{\text{list}}^{2}[i_{i-1}]$

    $\mathcal{A}\longleftarrow \mathtt{exp}((f_{i}^{2} \cdot f_{i-1}^{2})/\tau)$

    $\mathcal{A}_{\mathcal{N}}\longleftarrow \mathcal{A} \cdot \mathcal{N}$

    $m_{i} \longleftarrow \mathcal{A}_{\mathcal{N}} \cdot m_{i-1}$

    $m_{i} \longleftarrow \mathtt{argmax}(m_{i})$

    $m_{\text{list}} \longleftarrow \{m_{\text{list}}, m_i\}$

    }
\end{algorithm}
\end{minipage}
\end{center}
\vspace{+0.3cm}

The core of our object mask tracking from SD features is inspired from and analogous to {\textbf{(Q, K, V)} attention} \cite{vaswani2017attention}. Given the first frame GT mask $m_1$, analogically we assume $m_1$ and the next frame mask (\ie, $m_2$ in this case) to be comparable to {Key} (K) and {Query} (Q) of attention \cite{vaswani2017attention} respectively. Whereas, we consider the {affinity} between SD features of frame 1 and 2 to be the {Value} (V). Consequently, we hypothesise that the self-similarity between the first frame GT mask (K) and affinity matrix (V) would yield the segmentation mask of the immediate next frame (Q). Firstly, we extract SD UNet feature (from $\mathcal{U}_{\mathbf{d}}^{2}$) of all $N$ frames of a video. Next, we create a spatial neighbourhood mask (with a window $n$) via $\mathtt{SpatialMask}(\cdot)$ function to restrict the feature-affinity matrix within a \textit{local} spatial region. In practice, for every valid coordinate within the window $n$, $\mathtt{SpatialMask}(\cdot)$ makes the corresponding mask entry in $\mathcal{N}$ to be $1$. Astute readers may already have figured this to be similar to {local-attention} \cite{vaswani2017attention}. Starting from $i_2$, for each of the $N$ frames, we calculate the affinity matrix $\mathcal{A}$ between the current frame feature $f_{i}^2$ and its immediate predecessor $f_{i-1}^2$ as: $\mathcal{A}$=$\mathtt{exp}((f_{i}^{2} \cdot f_{i-1}^{2})/\tau)$. Where, the temperature hyperparameter $\tau$ is empirically set to be $0.2$. Next, we generate the neighbourhood-restricted feature-affinity matrix $\mathcal{A}_{\mathcal{N}}$ as: $\mathcal{A}_{\mathcal{N}}$=($\mathcal{A}\cdot\mathcal{N}$). We get the $i^{th}$ frame mask (starting from $i$=$2$ onwards) by multiplying $\mathcal{A}_{\mathcal{N}}$ with the $(i-1)^{th}$ frame mask\footnote{Either given (\ie, for $i=1$) or previously predicted (\ie, for $i=[2,3,4, \cdots, N]$).}. Eventually, we generate the final $i^{th}$ segmentation mask by applying $\mathtt{argmax}$ on $m_i$. An algorithm of this module is shown in Algo.~\ref{algo:seg}.

The research question remains as to how we make the predictions temporally-consistent \cite{grammatikopoulou2024spatio}. To this end, we resort to an efficient implementation trick. Instead of calculating the $i^{th}$ mask $m_i$ from the affinity with its \textit{immediate predecessor} (\ie, $m_{i-1}$), we accumulate the predicted masks at each step in a queue and utilise {all} these previously predicted masks to compute the affinity. Consequently, the value of $m_i$ is influenced by all preceding masks. This mechanism ensures that the algorithm considers a {limited history} of segmentation masks from past frames, thereby maintaining some level of temporal-consistency. Notably, being an online tracking method, our approach only relies on \textit{past frames} and \textit{previously predicted masks} to calculate the current mask, without utilising future frames.

\section{Experimental Results}
\label{sec:exp}
\keypoint{Dataset and Evaluation Metric.\ }We evaluate our method on the validation set of the CholecSeg8K \cite{hong2020cholecseg8k} dataset. CholecSeg8K contains segmentation masks for $8080$ frames from a total of $101$ (train:validation = $92$:$9$) clips of $17$ different laparoscopic cholecystectomy procedures operated by $13$ surgeons. The frames are annotated with segmentation masks, including $13$ distinct classes from three categories (\ie, $10$ organs, $2$ instruments, and background). Organs include the hepatic vein, liver, gastrointestinal tract, gallbladder, liver ligament, connective tissue, blood, cystic duct, abdominal wall, and fat. Whereas, the annotated instruments are l-hook electrocautery and grasper. Please note, we \textit{do not} use these GT masks for training. Following existing literature \cite{zhang2024towards, grammatikopoulou2024spatio, cheng2023segment}, we use mean Jaccard Score (\textbf{$\mathcal{J}_m$}), mean F-Score (\textbf{$\mathcal{F}_m$}), and per-pixel classification accuracy ($\mathcal{P}_{Acc.}$) as our primary evaluation metrics.

\keypoint{Implementation Details.\ }In all experiments we use Stable Diffusion \cite{rombach2022high} version $2.1$ and CLIP \cite{radford2021learning} textual encoder with embedding dimension of $1024$. Please note, we {do not} use any additional textual prompt. Instead, a {null-prompt} is used for all frames. We extract SD features from the $3^{rd}$ decoder of the UNet with timestep $t=200$. The size of $\mathtt{SpatialMask}(\cdot)$ window $n=50$. We incorporate the features and predicted masks from last $10$ frames for each mask prediction. All these parameters are ablated thoroughly in \cref{sec:abal}. Our inference pipeline takes only $\sim$$10$GB of VRAM and work in $0.5$ FPS, thus making it efficient for any consumer-grade GPU.

\keypoint{Competitors.\ }We compare with multiple self-designed baselines due to the lack of pre-trained internal representation-based temporal object tracking frameworks for surgical videos. We form our baselines with pre-trained models from {four} different paradigms. \textit{$\bullet$ Supervised \textbf{B}aseline:} In this category \textbf{B-ViT} and \textbf{B-ResNet50} uses features from ImageNet pre-trained ViT-S/16 \cite{dosovitskiy2021image} and Resnet50 backbones respectively. \textit{$\bullet$  Self-supervised \textbf{B}aseline:} In this class, \textbf{B-DINO}, \textbf{B-DINOv2}, and \textbf{B-MAE} utilises features from ViT-S/8 \cite{dosovitskiy2021image}, ViT-S/14 \cite{dosovitskiy2021image} and ViT-L/16 \cite{dosovitskiy2021image} backbones respectively, pre-trained with different self-supervision tasks. Ideally, these backbones should also possess object localisation to some extent, as their pre-training forces them to learn the same representation from different augmented versions of the same image \cite{oquab2023dinov2, caron2021emerging}. \textit{$\bullet$ Vision-language \textbf{B}aseline:} Here, \textbf{B-CLIP} and \textbf{B-MaskCLIP} extracts features from ViT-B/16 \cite{dosovitskiy2021image} backbone, pre-trained the large-scale image-text dataset with contrastive loss \cite{radford2021learning}. \textit{$\bullet$ Generative \textbf{B}aseline:} For a fair comparison, we also include a generative feature extraction method \textbf{B-SDXL}, which akin to ours, extracts features from a pre-trained Stable Diffusion XL Base-v1.0 model. Apart from the feature extraction backbone, all these baselines use the same segmentation module (\cref{sec:seg}) as ours. For all baselines, we extract features from the penultimate layer. To keep up with the state-of-the-art, we also compare with \textbf{SAM-Track} \cite{cheng2023segment}, which uses the popular Segment Anything Model (SAM) \cite{kirillov2023segment} and the Associating Objects with Transformers (AOT)-based tracker to track masks temporally. We also test this method by replacing the SAM backbone with medical image specialised MedSAM \cite{ma2024segment}. For the sake of completeness, we also compare with \textit{fully-supervised} methods like \textbf{SP-TCN} \cite{grammatikopoulou2024spatio}, and \textbf{HRNet32} \cite{sun2019high}. Numbers for these supervised methods are taken from \cite{grammatikopoulou2024spatio}. We evaluate all baselines in the same setup as ours.

\subsection{Performance Analysis}
\begin{figure}[!htbp]
    \centering
    \includegraphics[width=1\linewidth]{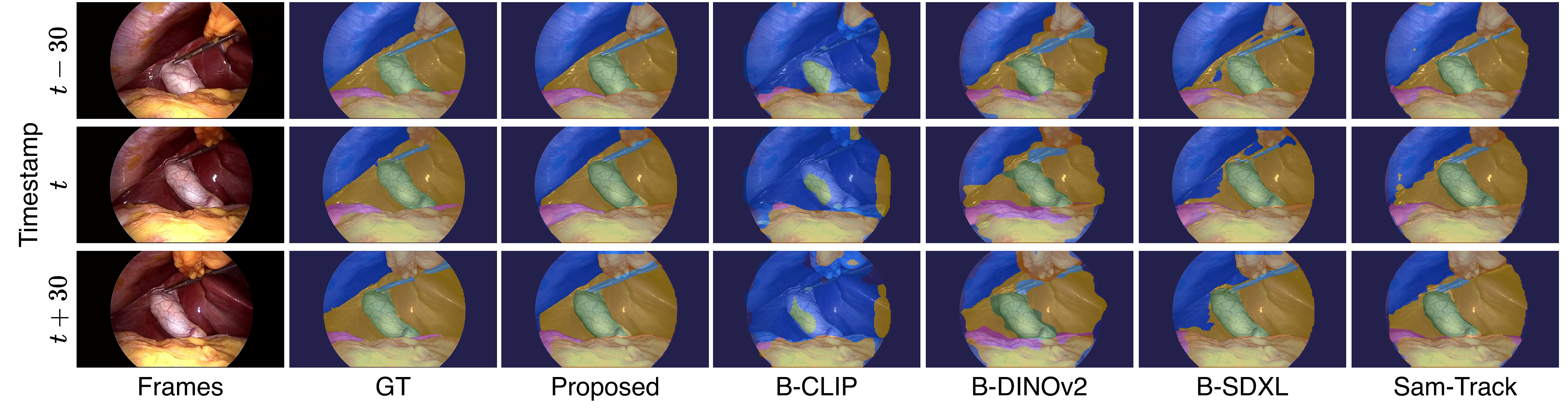}
    \caption{Qualitative comparison on CholecSeg8K \cite{hong2020cholecseg8k}. Notably, unlike the baseline competitors, the proposed method precisely tracks both smaller and larger anatomies across the temporal direction.}
    \label{fig:qual}
\end{figure}

\begin{figure}[!htbp]
    \centering
    \includegraphics[width=1\linewidth]{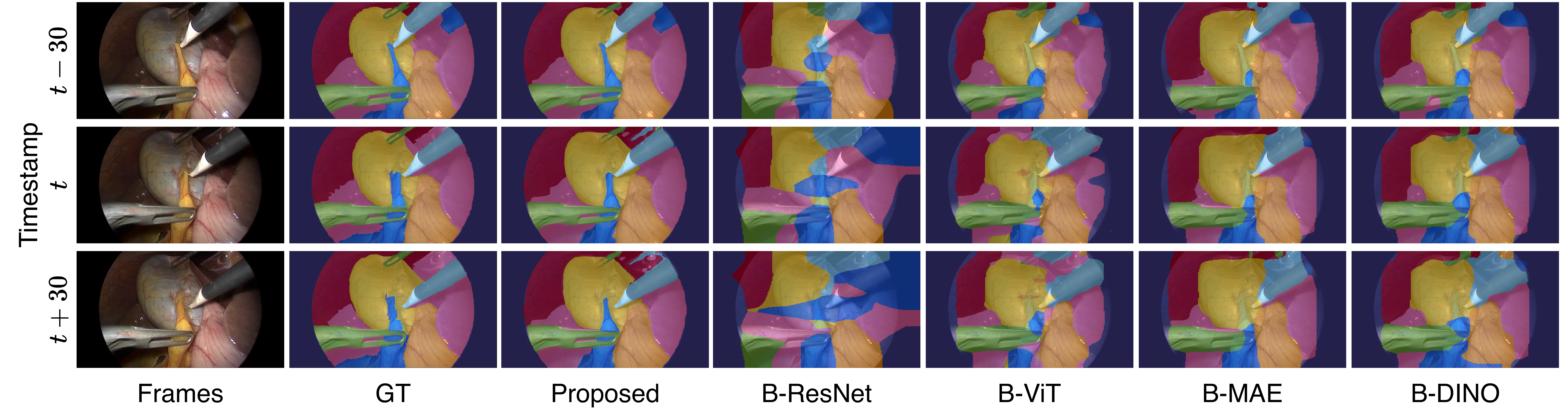}
    \caption{Qualitative comparison on CholecSeg8K \cite{hong2020cholecseg8k}. Noticeably, the proposed method surpasses competitors, maintaining accuracy even during rapid temporal movements of instruments and anatomis.}
    \label{fig:qual_1}
\end{figure}

To justify the usage of pre-trained SD as a feature extractor backbone for the task of temporal object tracking, we perform qualitative (\cref{fig:qual}-\ref{fig:qual_1}) and quantitative (\cref{tab:quant}) studies that demonstrate the architecture's latent potential to extract semantic information. Overall the proposed method outperforms all training-free methods with an average $\mathcal{J}_m$, $\mathcal{F}_m$ and $\mathcal{P}_{Acc.}$ gain of $13.80\%$ $8.61\%$ and $8.62\%$ respectively. 

\begin{table}[!htbp]
\setlength{\tabcolsep}{16pt}
\renewcommand{\arraystretch}{1}
\small
    \centering
    \begin{tabular}{llccc}
    \toprule
        \textbf{Methods} & \textbf{Backbone} & $\mathcal{P}_{Acc.}$ & $\mathcal{J}_m$ & $\mathcal{F}_m$ \\ \cmidrule(lr){1-5}
        B-MaskCLIP & ViT-B/16 & 69.10 & 39.26 & 69.43\\
        B-ResNet & ResNet50 & 54.08 & 27.83 & 54.29\\
        B-CLIP & ViT-B/16 & 64.91 & 31.68 & 65.29\\
        B-ViT & ViT-B/16 & 71.85 & 41.58 & 72.14\\
        B-DINO & ViT-S/8 & 77.08 & 52.14 & 77.37\\
        B-DINO & ViT-S/16 & 74.24 & 46.32 & 74.54\\
        B-DINOv2 & ViT-S/14 & 74.50 & 47.83 & 74.80\\
        B-MAE & ViT-L/16 & 72.26 & 43.51 & 72.56\\
        B-SDXL & SDXLv1.0 & 77.19 & 51.09 & 77.49\\
        \rowcolor{YellowGreen!40}
        \textbf{\textit{Proposed}} & SDv2.1 & {\textbf{79.19}} & {\textbf{56.20}} & \textbf{79.48}\\ \cmidrule(lr){1-5}
        SAM-Track \cite{cheng2023segment} & SAM & 78.74 & 55.59 & 78.48\\
        MedSAM-Track & MedSAM & 78.91 & 56.02 & 78.54\\
        HRNet32 \cite{sun2019high}$^\dag$ & HRNet32 & -- & 61.10 & --\\
        SP-TCN \cite{grammatikopoulou2024spatio}$^\dag$ & HRNet32 & -- & {65.37} & --\\
        \bottomrule
    \end{tabular}
    \caption{Quantitative comparison on CholecSeg8K \cite{hong2020cholecseg8k}. $\dag$ denotes {supervised} methods.}
    \label{tab:quant}
\end{table}

\cref{tab:quant}, and \cref{fig:qual}-\ref{fig:qual_1} shows the following observations -- \textit{(i)} the proposed method shows a {$24.48\%$} $\mathcal{J}_m$ gain over the vision-language baseline of \textbf{B-CLIP}. We posit that the CLIP \cite{radford2021learning} model being trained with vision-language contrastive loss, lacks the \textit{fine-grained} object-localisation capability required for dense prediction tasks such as object tracking (as seen in \cref{fig:qual}-\ref{fig:qual_1}). \textit{(ii)} Although \textbf{B-ViT} depicts better $\mathcal{P}_{Acc.}$ than \textbf{B-ResNet} due to its much larger backbone, it achieves {$14.58\%$} less $\mathcal{J}_m$ compared to the proposed method. We hypothesise this high-performance to be attributable to SD's \cite{rombach2022high} innate ability of \textit{semantic information retention} within its internal layers \cite{tang2023emergent}, gained during training. \textit{(iii)} Our method surpasses \textbf{B-DINOv2} with a $\mathcal{J}_m$ margin of {$8.33\%$}. We posit that the text-to-image diffusion pre-training with $l_2$ loss forces the SD model to generate {representative} feature for every pixel. Contrarily, DINO \cite{caron2021emerging} or CLIP \cite{radford2021learning} with their {image-level} loss objective fails to encompass enough representative features for dense prediction tasks (\cref{fig:qual}-\ref{fig:qual_1}). \textit{(iv)} While most baselines produce reasonable masks for {larger} organs and anatomies, they struggle in case of {finer structures} (\cref{fig:qual}-\ref{fig:qual_1}). We hypothesise that SD \cite{rombach2022high} excels at segmenting finer structures due to its Feature Pyramid Network \cite{lin2017feature}-like design, which has previously demonstrated marked improvement as a generic feature extractor in various dense prediction tasks \cite{lin2017feature}. \textit{(v)} \textbf{SAM-Track} despite using a much larger pre-trained model (\ie, SAM \cite{kirillov2023segment}), depicts less $\mathcal{J}_m$, $\mathcal{F}_m$, and $\mathcal{P}_{Acc.}$ than ours. While using medical specialised backbone (\ie, MedSAM \cite{ma2024segment}) offers some improvement, it remains inferior to the proposed method.  \textit{(vi)} Unsurprisingly, our \textit{training-free} method depicts slightly lower $\mathcal{J}_m$ than the temporal decoder-based \textit{fully-supervised} \textbf{SP-TCN} \cite{grammatikopoulou2024spatio} (\cref{tab:quant}). However, this indicates to a new avenue of training dedicated temporal decoders on top of the SD feature extractor, which we aim to explore in future.

Furthermore, to evaluate the generalisability of the proposed method, we conducted tests on two additional datasets: EndoVis-2015 \cite{bodenstedt2018comparative} and DAVIS-2017 \cite{pont20172017} containing surgical and non-surgical images respectively. In EndoVis-2015 (DAVIS-2017), our method achieves a $\mathcal{J}_m$ and $\mathcal{F}_m$ score of $83.81~(69.12)$ and $96.53~(73.21)$ respectively (\cref{tab:quant_new}), surpassing the training-free baselines with an average performance margin of $12.45\%$.

\begin{table}[!htbp]
\setlength{\tabcolsep}{20pt}
\renewcommand{\arraystretch}{1}
\small
    \centering
    \begin{tabular}{lcccc}
    \toprule
        \multirow{2}{*}{\textbf{Methods}}& \multicolumn{2}{c}{EndoVis-2015 \cite{bodenstedt2018comparative}} & \multicolumn{2}{c}{DAVIS-2017 \cite{pont20172017}} \\ \cmidrule(lr){2-3} \cmidrule(lr){4-5}
         & $\mathcal{J}_m$ & $\mathcal{F}_m$ & $\mathcal{J}_m$ & $\mathcal{F}_m$ \\ \cmidrule(lr){1-5}
        B-ResNet    & 56.61 & 73.05 & 40.25 & 44.92\\
        B-CLIP      & 63.32 & 83.34 & 45.82 & 61.23\\
        B-DINO      & 78.36 & 92.09 & 63.01 & 68.17\\
        B-DINOv2    & 75.16 & 90.62 & 57.73 & 64.03\\
        B-MAE       & 69.23 & 88.74 & 53.49 & 62.81\\
        B-SDXL      & 79.68 & 95.02 & 62.33 & 68.24\\
        \rowcolor{YellowGreen!40}
        \textbf{\textit{Proposed}} & {\textbf{83.81}} & {\textbf{96.53}} & \textbf{69.12} & \textbf{73.21}\\
        \bottomrule
    \end{tabular}
    \caption{Quantitative comparison on EndoVis-2015 \cite{bodenstedt2018comparative} and DAVIS-2017 \cite{pont20172017}.}
    \label{tab:quant_new}
\end{table}

\begin{figure}[!htbp]
    \centering
    \includegraphics[width=1\linewidth]{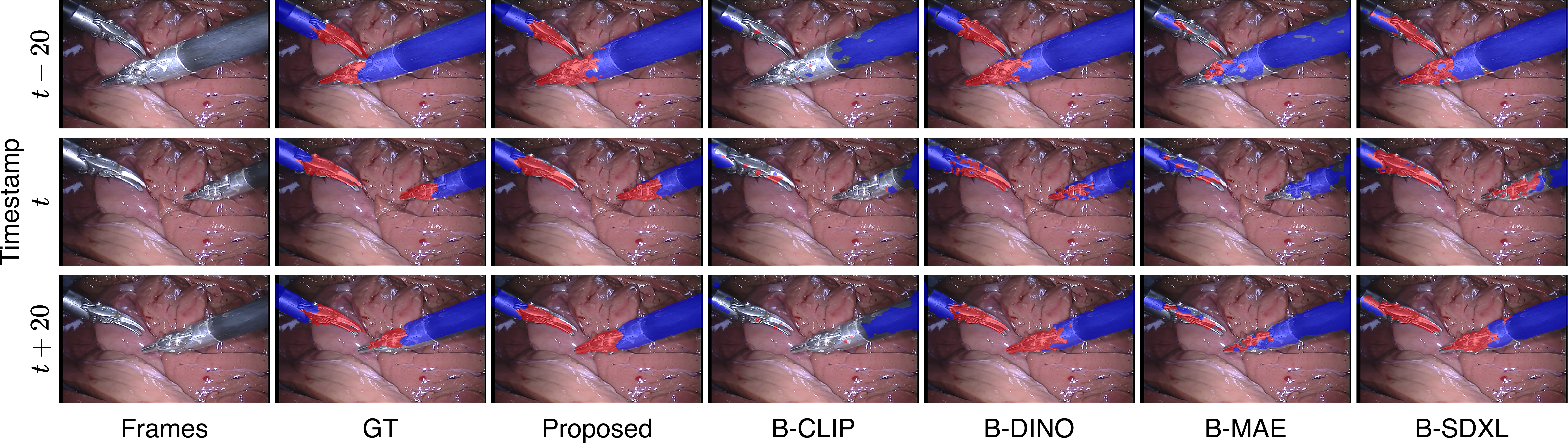}
    \caption{Qualitative comparison on EndoVis-2015 \cite{bodenstedt2018comparative}. Unlike baselines, the proposed method can track different parts (\eg, \red{$\blacksquare$}: Tip, \blue{$\blacksquare$}: Shaft) of the same instrument effectively.}
    \label{fig:qual_2}
\end{figure}

\subsection{Ablative Analysis}
\label{sec:abal}

\begin{figure}[!htbp]
    \centering
    \includegraphics[width=1\linewidth]{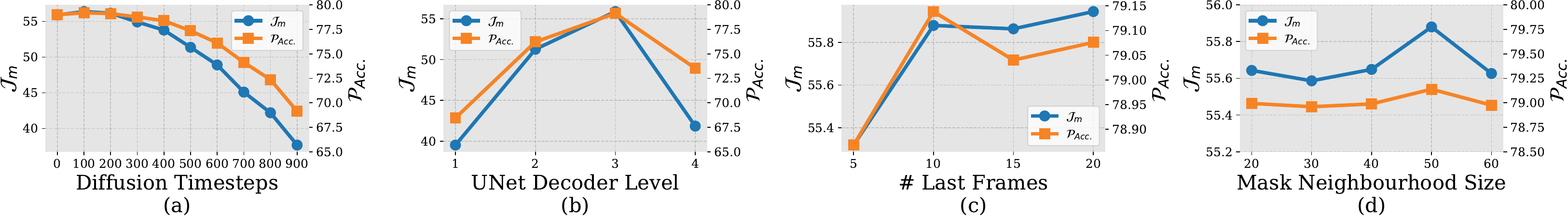}
    \caption{$\mathcal{P}_{Acc.}$ and $\mathcal{J}_m$ comparison on CholecSeg8K \cite{hong2020cholecseg8k} for different -- \textbf{(a)} diffusion timestep ($t$), \textbf{(b)} UNet \cite{ronneberger2015u} decoder $\mathcal{U}_{\mathbf{u}}^{n}$ level, \textbf{(c)} Number of previous frames considered for mask prediction, and \textbf{(d)} $\mathtt{SpatialMask}(\cdot)$ window size ($n$).}
    \label{fig:abal}
\end{figure}

\keypoint{\textit{[i]} Selecting the correct timestep.\ }The quality of the extracted SD feature depends significantly on the chosen timestep ($t$) (\cref{sec:diffusion}). Thus, selecting the optimum $t$ value is pivotal for task-specific adaptation of pre-trained SD models. Calculating $\mathcal{J}_m$ and $\mathcal{P}_{Acc.}$ on CholecSeg8K \cite{hong2020cholecseg8k} over varying $t$ reveals $t=200$ to be most optimal for temporal mask tracking. \cref{fig:abal}\blue{a} shows $\mathcal{J}_m$ to drop drastically with increasing $t$. We posit that, for larger $t$ values, the semantic structure of the input image distorts significantly, resulting in low $\mathcal{J}_m$.

\keypoint{\textit{[ii]} Are all decoder levels same?} As evident from \cref{fig:pilot}, different levels of the UNet decoder $\mathcal{U}_{\mathbf{u}}^{n}$ depicts different {feature-granularity} ($1\rightarrow4$ : coarse$\rightarrow$fine). Testing with different decoder features, we find $\mathcal{U}_{\mathbf{u}}^{3}$ features to yield the highest $\mathcal{J}_m$ (\cref{fig:abal}\blue{b}). While $\mathcal{U}_{\mathbf{u}}^{1}$ and $\mathcal{U}_{\mathbf{u}}^{2}$ features are too coarse for accurate object localisation, $\mathcal{U}_{\mathbf{u}}^{4}$ suffers with high-frequency noise (\cref{fig:pilot}). Whereas, $\mathcal{U}_{\mathbf{u}}^{3}$ provides a sweet-spot between coarse and fine-grained features.

\keypoint{\textit{[iii]} Effect of multi-frame features.\ }Our method considers a limited history of previous-masks for preserving temporal-consistency (\cref{sec:seg}). Experimentation with varying number of previous-predictions reveals that using $10$ past predictions yields the highest $\mathcal{J}_m$ (\cref{fig:abal}\blue{c}). We posit that employing a longer prediction history distorts the temporal information flow. Intuitively, this also explains why $\mathcal{P}_{Acc.}$ plummets when using more than $10$ past masks.

\keypoint{\textit{[iv]} Choice of mask window.\ }The $\mathtt{SpatialMask}(\cdot)$ function restricts the feature-affinity matrix within a local spatial region. Using a larger window size ($n$) often introduces irrelevant features in the affinity matrix, while small values of $n$ lead to loss of spatial context. We find $n=50$ to provide the highest $\mathcal{J}_m$ and $\mathcal{P}_{Acc.}$ (\cref{fig:abal}\blue{d}).

\begin{table}[!htbp]
\setlength{\tabcolsep}{15pt}
\renewcommand{\arraystretch}{1.5}
\small
    \centering
        \begin{tabular}{lccc}
    \toprule
        \textbf{SD Versions} & $\mathcal{P}_{Acc.}$ & $\mathcal{J}_m$ & $\mathcal{F}_m$\\
        \cmidrule(lr){1-4}
         v1.4 & 79.02 & 55.37 & 79.32\\
         v1.5 & 79.06 & 55.46 & 79.35\\
         v2.0 & 79.05 & 55.78 & 79.34\\
         \rowcolor{YellowGreen!40}
         \textbf{{Ours (v2.1)}} & \textbf{{79.19}} & \textbf{{56.20}} & \textbf{79.48}\\ \bottomrule
    \end{tabular}
    \caption{Ablating different SD \cite{rombach2022high} versions.}
    \label{tab:abal}
\end{table}

\keypoint{\textit{[v]} Effect of different SD versions.} We also compare different pre-trained versions of SD \cite{rombach2022high}. Although all SD version performs much better than other baselines, $\mathcal{J}_m$ and $\mathcal{F}_m$ numbers from v2.x models slightly outweigh the same from v1.x models. This behaviour is probably attributable to SD 2.x model's usage of much larger-scale OpenCLIP \cite{cherti2023reproducible} textual encoder during pre-training.

\subsection{Limitations and Future Works}
Although our method yields promising results (\cref{fig:qual}-\ref{fig:qual_1}) without any training or fine-tuning when given an accurate first frame mask, it sometimes struggles to maintain precise mask tracking for frames further along in the video. Nonetheless, even for slightly imperfect predictions from our method, human annotators would have a rational starting point to further annotate/correct upon, significantly lowering the cost of annotation from scratch. Moreover, in future we aim to train dedicated temporal decoders \cite{grammatikopoulou2024spatio} on top of the extracted SD features, which might further minimise imperfect predictions. Moving forward, we also aim to use pre-trained diffusion model features for other surgical video analysis tasks like phase recognition, keypoint tracking or depth-estimation. Also, our future works include runtime-reduction and \textit{fully-automated} tracking, thereby removing the need for a first-frame GT mask.

\section{Conclusion}
In conclusion, we have propose a training-free online framework for temporal object tracking in LC surgical videos, addressing the need for accurate anatomy and instrument localisation. By harnessing the latent potential of pre-trained text-to-image diffusion models and integrating cross-frame interactions, our method achieves impressive results surpassing similar state-of-the-art and baseline methods. Our approach not only offers a cost-effective solution to the challenges of pixel-level annotation but also holds promise for enhancing surgical guidance and post-operative analysis. Moreover, we believe our method to be an
important step towards aligning with the community’s progress towards
single pre-trained foundation model with task-specific adaptation.\\

\noindent \textbf{Funding} This work was funded by Medtronic plc.\\

\noindent \textbf{Data, code and/or material availability} Privately held.

\section*{Declarations}
\noindent \textbf{Conflict of interest} Mr. Koley, Mr. Barbarisi, Dr. Kadkhodamohammadi, Dr. Luengo, and Prof. Stoyanov are employees of Medtronic plc.\\

\noindent \textbf{Ethical approval} This article does not contain any studies with human participants or animals performed by any of the authors.

\bibliography{egbib}
\end{document}